\documentclass[runningheads,francais]{llncs}
\usepackage{graphicx}
\usepackage{xcolor}

\begin{document}
\title{Simulation réaliste d'utilisateurs pour les systèmes d'information en Cyber Range}
\titlerunning{Simulation réaliste d'utilisateurs en Cyber Range}
%
\author{Alexandre Dey\inst{1,2} \and Benjamin Costé\inst{1} \and Éric Totel\inst{3} \and Adrien Bécue\inst{1}}
\authorrunning{Alexandre Dey}
%
\institute{Airbus CyberSecurity, Rennes, France\and
IRISA, Rennes, France \and Télécom SudParis, Paris, France\\
\email{alexandre.dey@airbus.com}\\
\email{benjamin.b.coste@airbus.com}\\
\email{eric.totel@telecom-sudparis.eu}\\
\email{adrien.becue@airbus.com}}
\maketitle              
\begin{abstract}

La génération d'activité utilisateur est un élément-clé autant pour la qualification des produits de supervision de sécurité que pour la crédibilité des environnements d'analyse de l'attaquant. Ce travail aborde la génération automatique d'une telle activité en instrumentant chaque poste utilisateur à l'aide d'un agent externe; lequel combine des méthodes déterministes et d'apprentissage profond, qui le rendent adaptable à différents environnements, sans pour autant dégrader ses performances. La préparation de scénarios de vie cohérents à l'échelle du SI est assistée par des modèles de génération de conversations et de documents crédibles. 

\keywords{Cyber range  \and génération de texte \and reconnaissance d'images \and simulation de vie \and jeux de données \and honeynet}
\end{abstract}
\section{Introduction}

Les avancées récentes et la démocratisation des technologies de virtualisation (e.g., \textit{Software Defined Network} ou SDN, Cloud) ont notamment permis l'essor d'outils dédiés au cyber-entraînement. Appelées communément Cyber Ranges, ces plateformes facilitent le déploiement de Systèmes d'Information (SI) complets pour l'organisation de formations et exercices à destination des opérationnels de cyberdéfense. Outre leurs qualités pédagogiques, elles constituent également une base solide pour l'évaluation et la mise au point des outils de supervision de sécurité~\cite{pandora} (sonde de détection d'intrusion, détection d'anomalies, SIEM, etc.), ainsi que la constitution d'environnements d'analyse des attaquants (\textit{honeynet} et plateformes de détonation).

La simulation d'activité sur les postes utilisateur apporte une crédibilité nécessaire aux plateformes d'analyse des attaquants tout en permettant d'évaluer le comportement des produits de sécurité face au fonctionnement nominal des SI. En effet, les jeux de données disponibles pour l'apprentissage machine dans le domaine de la cybersécurité représentent des attaques, plus ou moins réalistes, mais n'intègrent pas ou peu de comportement illégitimes (scan de ports, branchement de clés USB sur des postes sensibles, etc.) émanant d'activités d'utilisateur légitimes. Cette absence d'activité, pourtant foisonnante sur un SI réel, complexifie l'immersion dans le cas du cyber-entraînement, biaise les données collectées sur les terminaux pour l'entraînement de méthodes de supervision basées sur l'apprentissage machine, et diminue grandement la crédibilité des plateformes d'analyse des attaquants.

La génération automatique de vie réaliste, objet notamment du challenge IA \& Cyber 2020 de l'ECW, est un sujet complexe qui demande de résoudre plusieurs problèmes. Tout d'abord, l'instrumentation des machines doit se faire par des méthodes extérieures à celles-ci afin de limiter les traces de simulation laissées sur les postes utilisateurs. En second lieu, il est nécessaire d'adapter en temps réel le scénario pré-établi aux réactions aléatoires de l'environnement de simulation (e.g., position d'une fenêtre, arrêt imprévu). Cette adaptation se fait via des vérifications automatiques de l'environnement qui doivent par ailleurs être effectuées en un temps inférieur ou égal au temps de réaction humain. Enfin, une assistance à l'opérateur est nécessaire pour la mise au point de scénarios à grande échelle.

Ce travail s'inspire des méthodes de génération de vie, reposant sur un agent, employées par la plateforme de détonation BEEZH, présentée par Amossys à la conférence C\&ESAR 2020~\cite{beezh}, pour lesquelles nous proposons plusieurs améliorations~:
\begin{itemize}
    \item Découpage de l'agent en plusieurs couches d'abstraction successives pour gagner en modularité (e.g., s'affranchir de la technologie de virtualisation, adaptabilité à des machines physiques, etc.);
    \item Combinaison efficiente de méthodes déterministes et d'apprentissage profond pour la réaction en temps réel de l'agent;
    \item Assistance à la création d'actions exécutables par l'agent;
    \item Assistance à la création de scénarios de vie à partir des profils des utilisateurs simulés et de leurs interactions.
\end{itemize}

Dans la section \ref{sec:sota}, nous présentons l'état de l'art. Les sections \ref{sec:actions} et \ref{sec:scenarios} décriront les travaux réalisés. Nous conclurons en discutant notre approche en sections~\ref{sec:discussion} et \ref{sec:conclusion}.

\section{État de l'art}
\label{sec:sota}


De nombreuses études ont souligné l'importance de la simulation d'activité utilisateur pour l'étude des logiciels malveillants \cite{Bulazel2017,Afianian2018}. Ainsi, l'absence de plusieurs artefacts (e.g., présence de fichiers dans la corbeille, présence de cookies, historique de navigation) permet d'identifier les machines n'ayant jamais été utilisées \cite{Miramirkhani2017}. Certains logiciels malveillants contournent les systèmes de détection par le biais de \emph{tests de Turing inversés} tels que des mécanismes à retardement (scroll dans un document Word) ou la vérification de la vitesse d'utilisation de la souris \cite{Vashisht2014}. La simulation en temps réel d'actions utilisateurs permet cependant de mettre en évidence le comportement malveillants de ces logiciels.

L'utilisation automatisée de l'interface graphique est un sujet historiquement étudié pour le contrôle qualité des logiciels. Deux approches s'y distinguent~: le rejeu d'actions pré-enregistrées~\cite{sikuli,singhera2008graphical,nguyen2014guitar}, long à configurer, et l'automatisation complète qui souffre d'un manque de réalisme~\cite{testar,Feng2020}. Chacune de ces méthodes requiert un agent installé sur les machines instrumentées, ce qui fournit un indicateur à l'attaquant, et teinte les journaux d'événements dans le cas de la collecte de jeux de données.

Plus récemment, MORRIGU \cite{Mills2021} instrumente des machines virtuelles via l'API VirtualBox depuis un hôte Windows. Malgré ses résultats positifs sur la détection de comportements malveillants, le manque de portabilité de la solution ne permet pas d'envisager son emploi à des fins de cyber-entraînement qui implique l'instrumentation de multiples machines en réseau. 

L'outil de simulation des utilisateurs intégré à la plateforme BEEZH~\cite{beezh} instrumente des machines virtuelles au travers de la fonction de déport d'écran de l'hyperviseur, reposant sur la technologie VNC. Les scénarios de vie sont découpés en actions unitaires simples (e.g., ouvrir le navigateur, recherche web) dont l'exécution est assistée par des méthodes d'analyse d'images (captures d'écran) à l'aide de la librairie desker~\cite{desker}, publiée sous licence GPL v3. Cette approche pionnière a montré l'importance mais également la complexité du sujet. Le recours à VNC offre en effet un large éventail de possibilités vis-à-vis de l'interaction avec la machine instrumentée mais se restreint néanmoins à l'instrumentation de machines virtuelles. La création des scénarios et leur adaptation aux spécificités de l'environnement nécessitent de surcroît des actions entièrement manuelles. Enfin, la solution d'analyse d'image retenue (Faster-RCNN~\cite{fasterrcnn}) sollicite d'importantes ressources de calcul incompatibles avec notre contrainte de performance. 
Faster-RCNN~\cite{fasterrcnn} est un algorithme très répandu (car pertinent) pour la détection de zones d'intérêt. D'autres initiatives plus récentes telles que RetinaNet~\cite{retinanet} ou SSD~\cite{ssd} fournissent cependant de meilleurs résultats en un temps plus court. 

La génération automatique de texte repose fréquemment sur les modèles de Markov cachés~\cite{hmm} (HMM, \textit{Hidden Markov Model}) mais les résultats produits deviennent rapidement incohérents lorsque la séquence générée s'allonge. Des solutions à base de réseaux de neurones comme Transformer~\cite{vaswani2017attention} apparaissent plus adaptés à la modélisation du langage. Les modèles basés sur le système GPT~\cite{GPT1}, performants et polyvalents, produisent toutefois des modèles massifs (e.g., 11 milliards de paramètres pour T5~\cite{T5}, 175 milliards pour GPT-3~\cite{GPT3}). Au regard de leurs performances, nos travaux s'inspirent de GPT-2~\cite{GPT2} (117 millions de paramètres) et CTRL~\cite{CTRL} pour la génération conditionnelle de texte.
 
\section{Exécution et enregistrement d'actions}
\label{sec:actions}

L'instrumentation de machines virtuelles ou physiques repose sur la conception d'un agent externe (Section \ref{ssec:agent}) capable de reconnaître son environnement et de s'y adapter en conséquence (Section \ref{ssec:img_proc}). La réalisation d'actions complexes pré-enregistrées (Section \ref{ssec:editor}) est également une fonctionnalité recherchée.

\subsection{Conception de l'agent}
\label{ssec:agent}

\begin{figure}[!ht]
	\center\includegraphics[width=\linewidth]{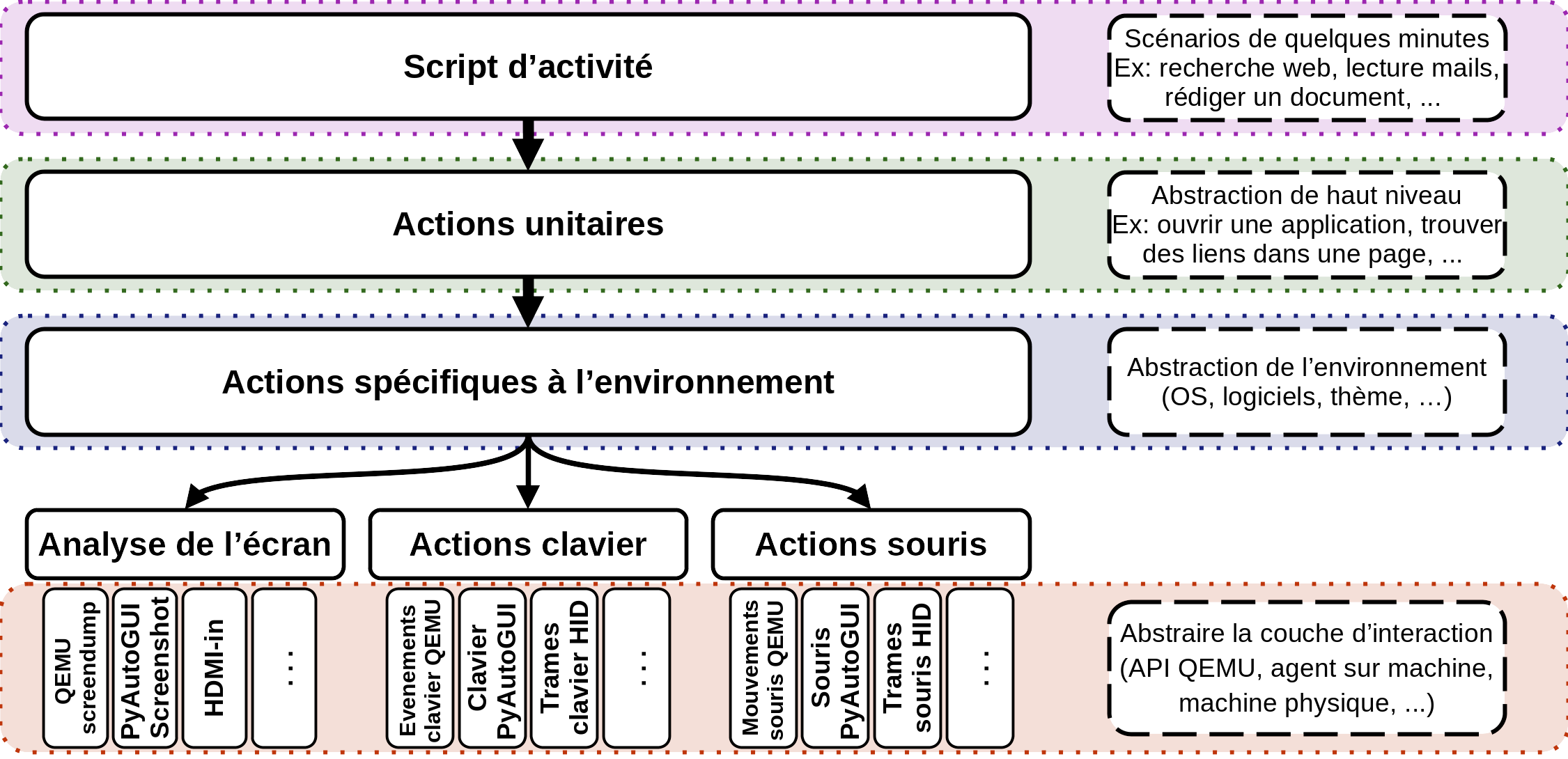}
	\caption{Architecture fonctionnelle de l'agent}
	\label{fig:archi_agent}
\end{figure}

Afin de simplifier la configuration du générateur de vie par les opérateurs, nous avons choisi une approche en couches d'abstraction (Fig.~\ref{fig:archi_agent}). En effet, une approche de bout-en-bout devient difficile à maintenir le nombre de scénarios grandissant (e.g., code dupliqué) et impose à l'opérateur de définir manuellement un grand nombre de détails lorsqu'il crée un scénario. Les scénarios d'activité utilisateur sont ainsi transcris par un agent en actions bas niveau (clavier, souris et écran) qu'il effectue ensuite sur la machine instrumentée. 

Au plus bas niveau, l'utilisateur virtuel interagit avec la machine instrumentée par le biais de la souris, du clavier et de l'écran. L'objectif est de pouvoir, sans modifier les scénarios de vie, adapter l'agent à plusieurs plateformes de virtualisation (e.g. VMWare ESXi, Proxmox, etc.) mais également à des machines physiques. Nous avons considéré pour cela trois méthodes d'interaction différentes mais complémentaires~: l'API Qemu Monitor\footnote{\url{https://qemu-project.gitlab.io/qemu/system/monitor.html}}, une connexion VNC et un agent installé sur la machine à instrumenter. Concernant les machines physiques, l'approche la plus pertinente semble la communication des actions clavier et souris par USB selon le protocole HID\footnote{\url{https://www.usb.org/hid}} en capturant la sortie vidéo.

Pour contrer des méthodes d'évasion de sandbox employées par certains logiciels malveillants, nous rajoutons de l'aléa dans les frappes clavier et les mouvements de souris. Pour ces derniers, chaque déplacement est découpé en une multitude de petits déplacements, et la vitesse varie en fonction de la distance à parcourir (i.e., un mouvement court sera moins rapide qu'on mouvement long). Ceci fluidifie le mouvement (augmentant le réalisme) et trompe les malware qui détectent les déplacements instantanés de la souris. De l'aléa et une inertie dans le mouvement sont rajoutés pour éviter les mouvements de souris rectilignes et trop précis (détectés par certaines méthodes d'évasion). 
Une latence aléatoire est également ajoutée entre chaque frappe clavier afin de rendre inopérantes les méthodes de détection recherchant une régularité. 
Enfin, lorsque l'utilisateur virtuel est en attente (e.g., lors de la lecture d'un texte, ou lorsque l'agent effectue un calcul long), des mouvements aléatoires de souris sont effectués pour simuler les mouvements spontanés lorsque la main est posée sur la souris. Les modèles d'aléa présentés ici visent à contrer les mécanismes de sécurité les plus simples. Ces modèles pourront être complexifiés par la suite afin de modéliser plus finement 
des utilisateurs humains (e.g., vitesse de frappe différente d'un utilisateur à un autre).  

La seconde couche dite d'interaction permet de s'adapter aux spécificités de l'environnement de la machine instrumentée. En effet, bien que les méthodes de reconnaissance d'images utilisées (section~\ref{ssec:img_proc}) soient peu sensibles aux variations graphiques mineures (e.g., résolutions d'écran différentes, couleurs légèrement différentes, etc.), certaines variations demandent de modifier les images à cibler. Par exemple, les actions à effectuer pour envoyer un mail avec Thunderbird ou Outlook sont fonctionnellement similaires mais les éléments d'interface sont suffisamment différents pour nécessiter des interactions différentes avec la machine.

La couche supérieure décrit des actions unitaires simples (e.g., ouvrir le navigateur web, rechercher un mot clé, identifier les liens dans du texte, etc.) et sert d'interface de programmation (API) pour les scénarios. Les scénarios ainsi créés constituent l'ultime couche d'abstraction.

\subsection{Analyse des captures d'écran}
\label{ssec:img_proc}

L'analyse des captures d'écran permet à un agent de contrôler son état en temps réel par reconnaissance des zones d'intérêts, telles les boutons d'interface à cliquer, les liens dans une page web, etc. Pour garantir l'efficience calculatoire, primordiale dans notre contexte, nous limitons systématiquement la quantité d'information (i.e. taille et nombre d'images) et la complexité des modèles statistiques. Trois techniques de reconnaissance d'image sont ainsi employées.
La première, connue sous le nom de \textit{template matching}, consiste à trouver dans une image (ici une capture d'écran) le ou les éléments correspondant le plus à une cible recherchée. Bien que de faibles variations (e.g., résolution, couleurs, etc.) suffisent à la rendre inopérante, cette méthode a l'avantage d'être peu coûteuse en ressources et d'être déterministe, ce qui la rend appropriée à la détection d'éléments d'interface qui varient peu voire pas pour un même environnement (e.g., bouton du menu démarrer, des fenêtres, etc.).

\begin{figure}[!ht]
	\center\includegraphics[width=0.9\linewidth]{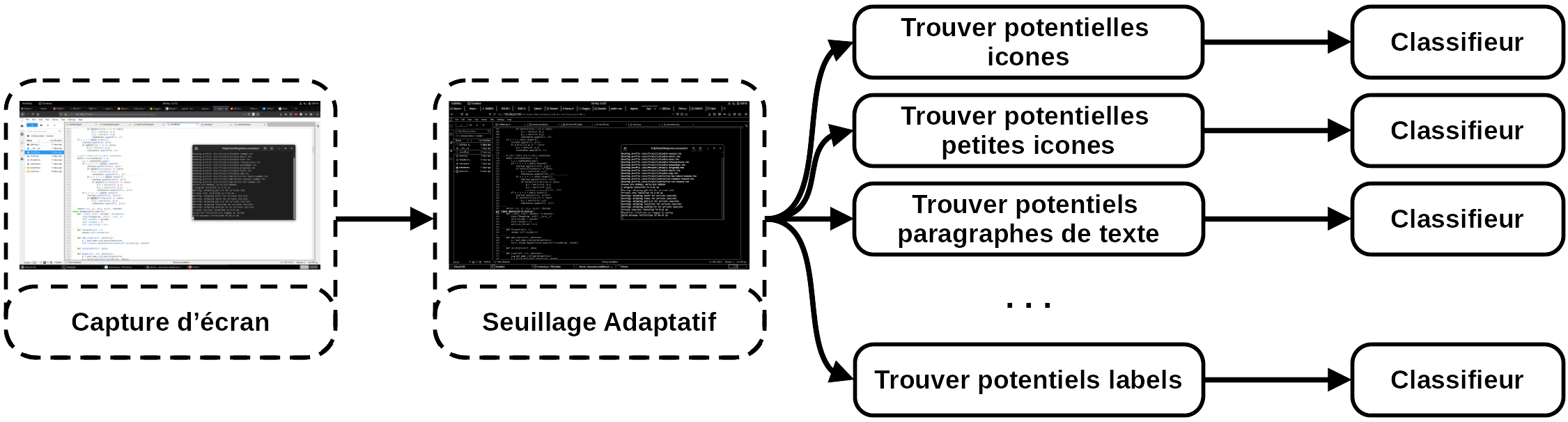}
	\caption{Méthodologie d'analyse d'images}
	\label{fig:image_analysis}
\end{figure}

Dans les cas où le \textit{template matching} n'est pas satisfaisant, nous proposons d'employer des algorithmes d'apprentissage profond ayant montré des performances remarquables pour la détection d'objet. Les méthodes les plus efficientes telles SSD~\cite{ssd} ou YOLO~\cite{yolo} requièrent cependant d'importantes ressources calculatoires qui limitent le passage à l'échelle (plusieurs machines instrumentées). De plus, il n'existe pas de jeux de données publics contenant des captures d'écran avec les zones d'intérêt détourées et annotées. Collecter et annoter un tel jeu de données prend un temps considérable. Pour ces raisons, nous avons choisi une approche qui exploite la géométrie des formes à repérer à l'écran (Fig.~\ref{fig:image_analysis}) en appliquant une méthode de seuillage adaptatif~\cite{thresh}. Les objets au premier plan (e.g., icônes, texte) ainsi que les séparations entre les différentes zones à l'écran (e.g., contour des fenêtres) deviennent ainsi simples à identifier. En appliquant des règles de filtrage (e.g., une icône est à peu près aussi large que haute, un bouton sera plutôt plus large que haut, etc.) il est possible, pour un coût calculatoire faible, d'identifier les zones d'intérêt. Des modèles statistiques (e.g., réseaux de neurones à convolution) filtrent les potentiels faux positifs et classifient les autres.   

Enfin, la reconnaissance de caractères est utilisée pour la navigation dans l'interface (e.g., label des fichiers et dossiers, éléments des menus, etc.) ainsi que la reconnaissance des liens dans les pages web. Notre implémentation utilise la librairie tesseract~\cite{tesseract} pour l'analyse des éléments textuels de l'interface, et reprend la même méthode que desker~\cite{desker} pour la détection de liens.

\subsection{Aide à la création d'actions}
\label{ssec:editor}

La reconnaissance d'image présentée dans la section~\ref{ssec:img_proc} donne à l'agent la capacité de reconnaître l'environnement dans lequel il évolue et s'y adapter. Par exemple, l'interface de Firefox diffère de celle de Chrome, bien que les fonctionnalités de ces navigateurs soient équivalentes. Par conséquent, l'agent cherchera des repères visuels différents pour les manipuler. Nous proposons un outil de création d'actions qui exploite ces similarités pour faciliter la collecte de données destinées aux techniques de reconnaissance d'images présentées plus haut.

Cet outil enregistre les mouvements et clics de la souris, les frappes du clavier ainsi que des captures d'écran. Ces données brutes sont ensuite agrégées pour découper l'enregistrement en petites actions (e.g., mouvement de la souris de A vers B, puis double clic gauche sur B). Les zones d'intérêts dans les captures d'écran sont extraites selon deux méthodes différentes. La première consiste à comparer une image prise au moment d'un clic de souris avec l'images prise au début du mouvement précédent ce clic. La plupart des interfaces graphiques actuelles mettent en surbrillance les éléments survolés par la souris ce qui permet d'extraire simplement ces éléments, qui serviront de cibles de \textit{pattern matching}. La seconde méthode consiste à employer la technique d'identification des zones potentiellement intéressantes décrite dans la section~\ref{ssec:img_proc} (Fig. \ref{fig:image_analysis}) ce qui accélère grandement l'annotation des jeux de données pour l'entraînement de modèles de reconnaissance de zones d'intérêt. 

\section{Gestion de scénarios de vie à l'échelle du système}
\label{sec:scenarios}

Les outils et méthodes présentées permettent à notre agent d'interagir et de s'adapter à son environnement. La présente section aborde la réalisation et l'exécution de scénarios de complexités variées à l'échelle d'un SI comportant plusieurs machines et utilisateurs.

\subsection{Orchestration des communications entre utilisateurs}
\label{ssec:orchestration}

Dans le contexte du cyber-entraînement, pour améliorer la cohérence et simplifier la mise en place de l'exercice, il est intéressant de créer des avatars pour chacun des utilisateurs simulés et de leurs interactions. En fonction de leurs rôles dans l'entreprise et des relations qu'ils entretiennent avec leurs collègues, les employés vont interagir différemment avec le système d'information simulé. Par exemple, deux collègues amis en dehors de leur travail sont plus susceptibles de discuter par mail de manière informelle que deux collègues ne se connaissant pas. Similairement, un développeur utilisera plus souvent l'éditeur de code qu'un manager.

\begin{figure}[!ht]
	\center\includegraphics[width=0.9\linewidth]{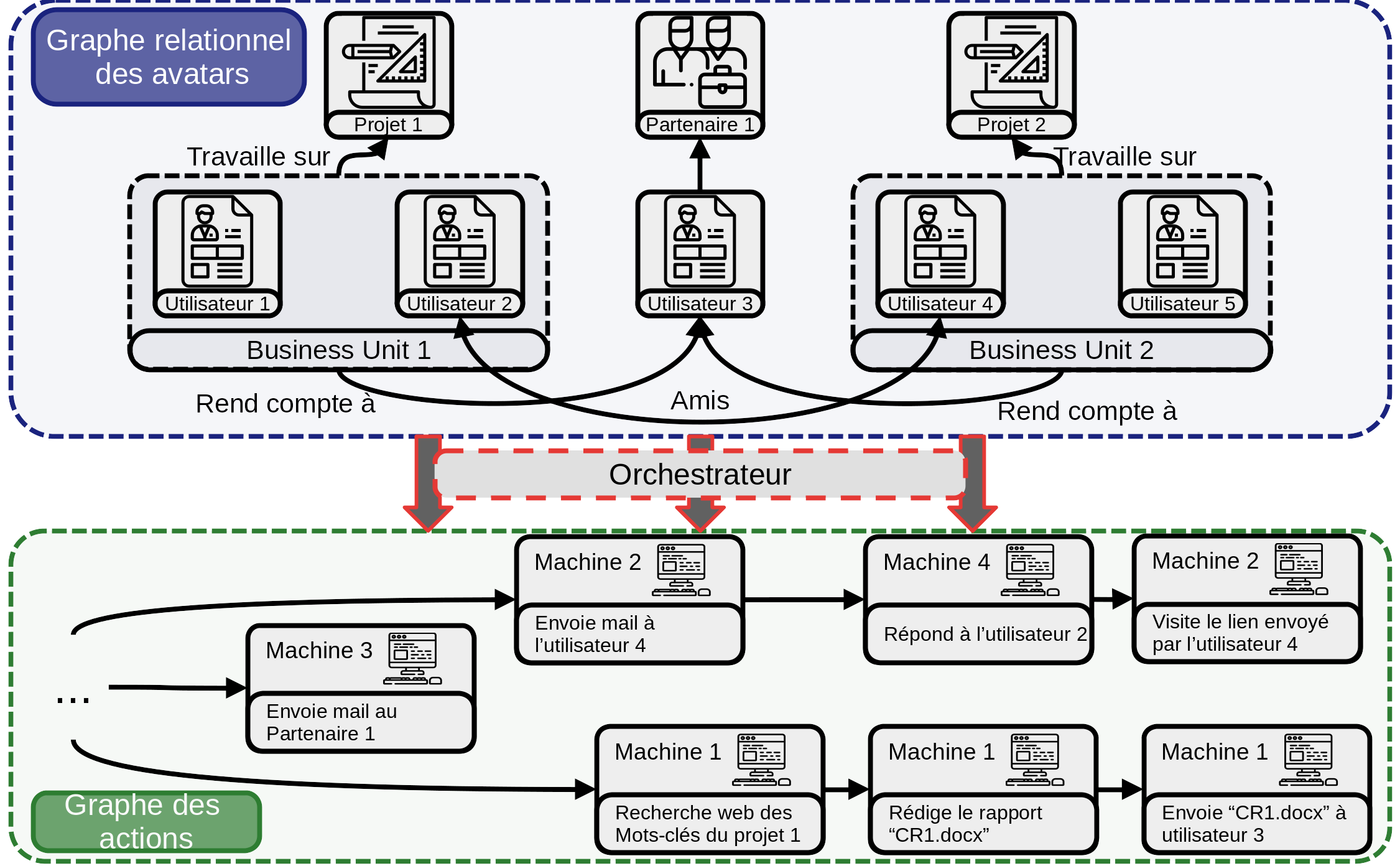}
	\caption{Vue d'ensemble de l'orchestrateur}
	\label{fig:scheduler}
\end{figure}

Nous générons des canevas de scénarios de vie à l'échelle du système par le biais d'un graphe relationnel des utilisateurs de ce système (Fig.~\ref{fig:scheduler}). Ce graphe est composé des différents avatars, des projets sur lesquels ils travaillent, de leurs groupes de travail et de la nature des relations qu'ils entretiennent entre eux (e.g., amicales, client-fournisseur, partenaires, lien hiérarchique, etc.).

\subsection{Modèles de génération de texte}
\label{ssec:text_gen}

L'ajout de contenu réaliste renforce notablement la crédibilité de la vie simulée. Notamment, l'absence d'échanges de mails ou de documents sur les postes utilisateur est un indicateur fort pour un adversaire, y compris si le contenu des mails et des documents est manifestement non crédible (e.g., succession de mots vide de sens). Cependant, la génération manuelle de ce contenu est laborieuse. En effet, rédiger du texte à la manière d'un avatar est en soi une tâche fastidieuse qu'il est pourtant nécessaire de reproduire pour plusieurs dizaines d'avatars dans le cas de scénarios complexes.

Le domaine du traitement du langage naturel (NLP, \textit{Natural Language Processing}) a récemment connu une avancée majeure avec la démocratisation des réseaux de neurones type Transformer~\cite{vaswani2017attention}. Le mécanisme d'attention, principale particularité de ces modèles, leur permet de modéliser avec précision la syntaxe et la sémantique de plusieurs langages naturels. OpenAI a démontré l'efficacité de l'apprentissage par transfert avec des modèles comme GPT~\cite{GPT1}, ainsi que ses capacités dans le domaine de la génération de texte. Le modèle employé génère du texte dans la continuité d'un contexte fourni au préalable (souvent des mots commençant une phrase). Dans notre cas, nous affinons le modèle pré-entraîné GPT-2~\cite{GPT2} pour la génération conditionnelle de texte. Plus spécifiquement, le contexte fourni en entrée du modèle contient des informations sur le ton à employer dans le texte (e.g., professionnel, informel, à un partenaire, à un client) ainsi que les thématiques à aborder (e.g., le sujet du mail, des mots clés).

\begin{figure}[!ht]
    \centering
    \includegraphics[width=\textwidth]{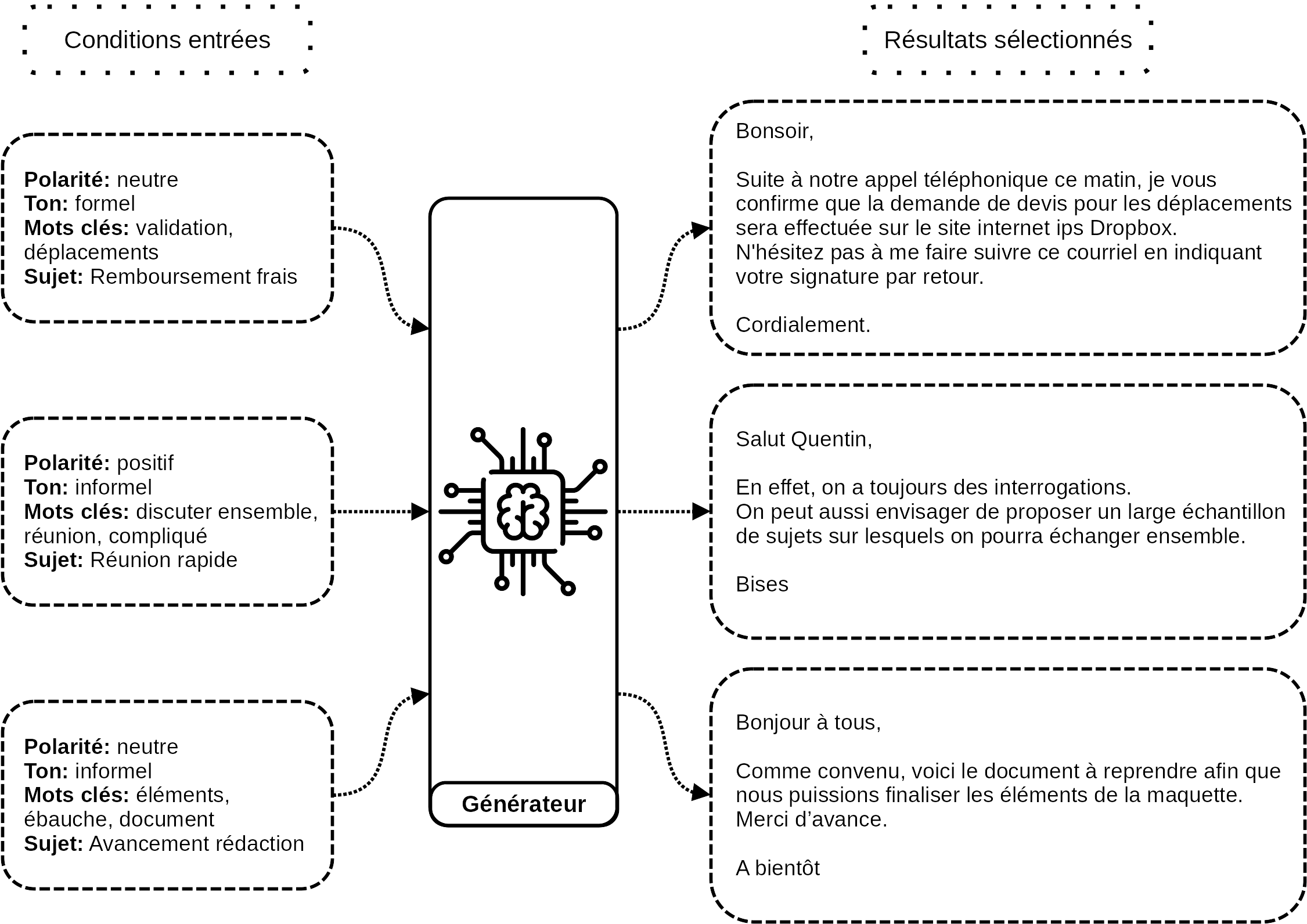}
    \caption{Exemple de mails générés conditionnellement avec le modèle dérivé de GPT-2}
    \label{fig:text_gen}
\end{figure}

Pour vérifier l'efficacité de notre méthode de génération de texte, nous avons entraîné un modèle sur un jeu de donnée de 35~000 mails récupérés en source ouverte. Ces mails sont associés à un ensemble de mots-clés, une polarité (positif, négatif ou neutre) et une tonalité (formel, informel). 

\section{Discussion}
\label{sec:discussion}

\subsection{Actions exécutées par l'agent}

Afin de ne pas laisser de traces d'instrumentation sur la machine cible, nous avons développé un agent qui communique au travers d'interfaces externes (clavier, souris, écran), le rendant ainsi indétectable pour un attaquant ou un outil de supervision (e.g., EDR). L'agent développé est apte à effectuer des activités de bureautique simples (lecture/écriture de mail, navigation web, traitement de texte, etc.). L'automatisation est de plus masquée par l'ajout d'un aléa dans les mouvements de souris et les frappes clavier qui compliquent la détection automatique de cette instrumentation (ce que font certains logiciels malveillants pour détecter des sandbox \cite{Vashisht2014}), obligeant ainsi un attaquant à évaluer la crédibilité du scénario en lui même.

Le panel des actions que l'agent peut réaliser améliore significativement la robustesse des systèmes actuels face à des malware utilisant des tests de Turing inversés. En revanche, la bibliothèque d'actions devra être complétée pour créer des jeux de données réalistes destinés à valider le fonctionnement d'algorithmes de détection. En particulier, la réalisation de tâches d'administration et plus largement des actions générant du bruit au niveau des outils de supervision (e.g., un administrateur ne respectant pas les procédures, un utilisateur utilisant des outils de scan réseau, etc.) permettraient de différencier les niveaux techniques des utilisateurs simulés et, par la même occasion, d'enrichir les jeux de données pour la détection. 

L'enregistreur d'actions permet d'adapter en quelques minutes les actions existantes à des environnements nouveaux. Ceci permet d'une part d'intégrer rapidement, sans connaissances préalables sur le fonctionnement de notre outil, de nouvelles actions supportant des cas non gérés (e.g., une nouvelle pop-up pour la gestion des cookies lors de la navigation web, une mise à jour majeure d'un logiciel, etc.) et d'autre part, de réaliser des actions complexes qui nécessiteraient un temps de développement important. L'adaptation de scénarios complets, bien que grandement facilitée, demeure toutefois une activité complexe.

\subsection{Génération de scénarios}
Ce travail présente la conception d'un agent seul et ne traite donc pas de l'orchestration des agents au sein d'un environnement multi-machines. Cette problématique comprend la communication et l'organisation entre les agents (e.g., répondre à mail) mais également la gestion de profils utilisateurs brièvement abordée dans la Section \ref{ssec:orchestration}. 

De plus, le modèle de génération de texte actuellement développé génère une majorité de résultats non crédibles~: environ 20\% sont exploitables avec des modifications mineures. Ceci nous empêche de générer la totalité des échanges sans intervention humaine. Bien que la génération ne soit pas entièrement automatique, celle-ci est toutefois grandement facilitée. En effet, il est possible, à partir du même contexte, de générer plusieurs candidats que l'opérateur peut utiliser pour gagner du temps dans la génération des mails et documents du scénario. Nous envisageons deux axes majeurs pour améliorer ces capacités de génération~:
\begin{enumerate}
    \item Augmenter la quantité de données et la taille du modèle (e.g., utiliser une autre variante de GPT-2);
    \item Rajouter une étape d'entraînement type GAN (Generative Adversarial Network) \cite{croce-etal-2020-gan}, pour inciter le modèle à générer des candidats plus réalistes.
\end{enumerate}

\section{Conclusion}
\label{sec:conclusion} 

Nous avons présenté une méthode pour simuler des utilisateurs à l'échelle d'un SI complet en favorisant l'adaptabilité et la simplicité de mise en \oe{}uvre. Notre approche emploie un agent externe découpé en plusieurs niveaux d'abstraction, avec, au plus bas, une interaction avec les machines instrumentées au travers du clavier, de la souris et de l'écran. Pour faire face à la diversité des interfaces utilisateur des systèmes modernes (e.g., OS différents, logiciels différents, etc.) l'agent embarque des techniques de reconnaissance d'images reposant à la fois sur des méthodes déterministes (template matching) et d'apprentissage profond (réseaux de neurones à convolution). Ceci équilibre la quantité de ressources calculatoires requises par l'agent avec la flexibilité offerte par les méthodes probabilistes. Un enregistreur d'actions simplifie la configuration de l'agent pour tout nouvel environnement en extrayant automatiquement les images cibles, et en facilitant la collecte et l'annotation de données d'entraînement pour les modèles statistiques. La création de scénarios de vie se base sur les avatars des utilisateurs virtuels et de leurs interactions dont se nourrissent nos modèles de génération conditionnelle de texte qui produisent en masse des conversations e-mail réalistes et des documents crédibles.

Notre proposition enrichit la méthode de génération de vie initialement proposée pour la plateforme BEEZH par l'amélioration des performances et le réalisme de l'activité générée. Nous complétons également la proposition initiale par une assistance pour la génération de scénarios à grande échelle. En cours d'implémentation, nos travaux bénéficient de premiers résultats encourageant qui restent toutefois à consolider avant de valider expérimentalement notre proposition.


\subsubsection*{Acknowledgments}

Les auteurs remercient chaleureusement Alexandre De Beaudrap dont les travaux de stage ont permis de consolider l'approche.

%
%
%
\bibliographystyle{splncs04}
\bibliography{biblio}

\end{document}